\documentclass[pmlr]{jmlr}

\usepackage{longtable}

\usepackage{booktabs}
\usepackage[load-configurations=version-1]{siunitx} 

\makeatletter
\def\set@curr@file#1{\def\@curr@file{#1}} 
\makeatother

\theorembodyfont{\upshape}
\theoremheaderfont{\scshape}
\theorempostheader{:}
\theoremsep{\newline}

\newcommand{\comment}[1]{}
\usepackage{graphicx}
\usepackage{lineno}
\usepackage{amsmath}
\usepackage{adjustbox}
\usepackage{float}
\usepackage{wrapfig}
\usepackage{lipsum}
\usepackage{enumitem}
\usepackage{multirow}

\title[Explainability of traditional \& deep learning models]{
Explainability of Traditional and Deep Learning Models on Longitudinal Healthcare Records}

\author{\Name{Lin Lee Cheong}
       \Email{lcheong@amazon.com}\\ 
       \addr Amazon ML Solutions Lab\\
       Palo Alto, CA, USA \\
       \AND
       \Name{Tesfagabir Meharizghi}
       \Email{mehariz@amazon.com}\\ 
       \addr Amazon ML Solutions Lab\\
       Palo Alto, CA, USA \\
       \AND
       \Name{Wynona Black}
       \Email{wynona.black@merck.com}\\ 
       \addr Merck \& Co., Inc. \\
       Kenilworth, NJ, USA \\
       \AND
       \Name{Yang Guang}
       \Email{yaguan@amazon.com}\\ 
       \addr Amazon ML Solutions Lab\\
       Palo Alto, CA, USA \\
       \AND
       \Name{Weilin Meng}
       \Email{weilin.meng@merck.com}\\ 
       \addr Merck \& Co., Inc. \\
       Kenilworth, NJ, USA} %

\begin{document}
\restylefloat{figure}
\maketitle

\begin{abstract}
  Recent advances in deep learning have led to interest in training deep learning models on longitudinal healthcare records to predict a range of medical events, 
  with models demonstrating high predictive performance. Predictive performance is necessary but insufficient, however, with
  explanations and reasoning from models required to convince clinicians for sustained use. Rigorous evaluation of explainability is often missing, as comparisons between models (traditional versus deep) and various explainability methods have not been well-studied. Furthermore, ground truths needed to evaluate explainability can be highly subjective depending on the clinician's perspective. Our work is one of the first to evaluate explainability performance between and within traditional (XGBoost) and deep learning (LSTM with Attention) models on both a global and individual per-prediction level on longitudinal healthcare data. We compared explainability using three popular methods: 1) SHapley Additive exPlanations (SHAP), 2) Layer-Wise Relevance Propagation (LRP), and 3) Attention. These implementations were applied on  synthetically generated datasets with designed ground-truths and a real-world medicare claims dataset. We showed that overall, LSTMs with SHAP or LRP provides superior explainability compared to XGBoost on both the global and local level, while LSTM with dot-product attention failed to produce reasonable ones. 
  With the explosion of the volume of healthcare data and deep learning progress, the need to evaluate explainability will be pivotal towards successful adoption of deep learning models in healthcare settings.
\end{abstract}

\section{Introduction}
Longitudinal healthcare records are defined as sets of health information generated temporally by one or more encounters in a care setting.
As the digitization of longitudinal healthcare records data continues, traditional 
modeling techniques are increasingly challenged in 
understanding the high-dimensional complexity and non-linear nature 
of medical data. Deep learning models, especially those originating 
from natural language processsing (NLP) field,
have found predictive success when applied to longitudinal healthcare data. 
These models are capable of leveraging 
time-dependent and complex data beyond typical 
human capabilities, as shown by \cite{icu_models} on raw ICU data with
sequence models.

Success in prediction performance is necessary but insufficient 
for adoption of AI-based models by the
healthcare community. \cite{what_clinicians_want} showed that 
high level of transparency from a ML-based system 
is necessary for sustained use
in clinical settings, and 
remains a significant barrier. \cite{icu_model_comment} 
highlighted this transparency difficulty on \cite{icu_models}'s work 
, `unlike models such as the simple logistic prediction model
... deep learning rests upon
hidden factors that uncover complex, high-order data patterns; by definition,
we do not know — and far less are able to describe or explain — what
such a model is doing.' Recent healthcare applications of 
deep learning continue to focus on predictive performance 
metrics such as area under curve (AUC), with model explainability receiving qualitative treatment. 
Examples include \cite{rajkomar} on predicting readmissions, 
mortality and length of stay with LSTMs with a single example of interpretability. \cite{zhang} 
qualitatively described attention-based LSTM model interpretation of 
two different patient cases for all-cause hospitalization. 
\cite{medbert} similarly notes medical concept patterns 
learned by a BERT-based model. Consequently, traditional ML methods 
continue to be used, including \cite{bos} and \cite{ross}.


It is thus of clinical relevance and importance that 
more studies are conducted by healthcare-focused ML
practitioners into understanding model behavior.
It is also currently unclear how explainability derived from more
traditional approaches
compare with quantitative explainability 
or interpretability derived from deep learning methods 
when applied to longitudinal healthcare data. This lack of quantitative focus 
on healthcare model explainability and lack
of comparisons between traditional ML and deep learning based approaches
is primarily due the varied approaches and expectations 
surrounding this topic. There is currently little consensus 
within the ML community, including healthcare-focused
ML researchers, on the definition of interpretability and explainability. Some
practitioners, including us in this work, 
have chosen to use both terms interchangeably, while
others including \cite{rudin} draw distinctions between the two. 
Secondly, there is also the \textit{type} of interpretability/explainability: 
global interpretability seeks to explain overall model behavior; 
local interpretability seeks to explain at the per prediction 
level (\cite{hc_interpretability}). Thirdly, there is the distinction of model-agnostic 
versus model-specific type of explainability. 
Model-agnostic approaches, including 
SHAP treat an already-trained model as a black box while providing interpretations. \cite{lundberg_shap} 
utilized SHAP to provide visual interpretations for a tree-based model on the
epidemiological data and for hypoxaemia during surgery (\cite{lundberg_hypox}). 

Distinct from model-agnostic approaches, 
model-specific approaches require understanding of the 
underlying model architecture. This includes Layer-wise Relevance Propagation (LRP),
a general method for explaining deep learning
models originally proposed by \cite{lrp}. LRP
provides explainability by backwards-attributing prediction scores
through the network back to the input space.
Other NLP researchers provide explainability 
via inclusion of attention mechanisms within the model 
architecture. The extracted attention scores is thought to 
provide insight into a deep learning model's mechanisms, 
with inputs or time-steps (for sequence models) 
with high attention scores believed to be of higher 
importance. This has very recently been called into question 
and is area of active debate, as evidenced by \cite{att_is_not_explaination}'s work
titled `Attention is not Explanation' and a responding paper by
\cite{att_is_not_not_explaination} titled  `Attention is not not 
Explanation'. 

Finally, the type of healthcare data studied heavily determines 
ML model and explainability. On the simpler end, 
traditional methods can
provide accurate and explainable predictions with a 
curated set of well-understood predictors.
On the other extreme,  temporally organized datasets such as
electronic health records (EHR) and claims datasets allow for 
complex deep learning approaches such as LSTM, BERT and Transformers. 
These models account for time-dependencies and 
complex feature interactions where traditional ML methods have struggled. 
We focus on medical claims datasets as claims datasets are better suited to capture longitudinal medical information,
and have found success using both traditional and deep learning models and thus well-suited for
a comparative study.

\subsection*{Contributions}
Model explainability is explored within the context of medical claims datasets. In this work, we investigate and quantify the 
classification and explainability performance using one traditional model (XGBoost), one deep learning model (LSTM) and 
three promising explainability methods: SHAP,
LRP and Attention. This provides four combinations to evaluate: SHAP-XGBoost, SHAP-LSTM, LRP-LSTM and Attention-LSTM.
We quantitatively show that SHAP-LSTM and LRP-LSTM provide reliable global and local interpretability, and Attention-LSTM 
(dot-product attention and self-attention variants) has sub-par explainability performance. 

We detail methods to create realistic synthetic datasets that
reflect the properties of medical claims datasets with a binary classification task, which sidesteps issues experienced by other researchers working in this area who primarily 
rely on non-representative datasets such as the MNIST digits dataset or difficult-to-evaluate real-world datasets.
Datasets, with and without temporal dependencies, provide fair evaluation to both
XGBoost and LSTM (which explicitly addresses temporality). We identify the importance of
using set similarity metrics to inform model training instead
of solely relying on predictive performance like AUC. 
This ensures that the trained model and explainability method
provides the optimal transparency to clinicians to inspire confidence, both on a global and local level. 
Finally, learnings are applied to a real-world medical claims dataset to predict \textit{C.Diff} infection.

\subsection*{Generalizable Insights about Machine Learning in the Context of Healthcare}
\begin{itemize}[topsep=0pt,itemsep=-1ex,partopsep=1ex,parsep=1ex]
    \item Deep learning models, used with SHAP or LRP for explainability, provides superior explainability in a medical claims setting. These combinations provide both global explainability (overall important risk factors or key drivers of prediction) and
    at per-patient (local) level. The explainabilities, relative to the SHAP-XGBoost, are generally consistent but slightly different.
    \item Attention with deep learning LSTM model does not provide satisfactory explainability when applied to healthcare, despite providing excellent predictive performance. Self-attention performed better than simple dot-product attention.
    \item Healthcare ML modelers must apply caution during model training, as models with too-large parameter space will provide similar predictive performance to a just-right model but with reduced explainability performance. 
    \item Our approach is helpful in improving understanding from machine learning models to inspire clinician confidence, and compare
            deep learning models with the historically better understood and better adopted traditional ML models. 
            The evaluated SHAP/LRP methods are general methods that can be broadly applied to many classes of deep learning models, and this work serves as confidence that the deep learning models with SHAP or LRP explainability can provide
            superior explainability to traditional approaches at both global and per patient (local) level.
\end{itemize}

\section{Related Work}
Additional related work that overlaps with our focus area are described. 
We did not find papers that wholly describe
our combined area of focus: using medical claims dataset, to study a combination of 
deep learning/traditional machine learning models with model-agnostic and
model-specific explainability, and utilizing ground-truth or clear cut synthetic datasets
to understand the quality and assumptions of explainability. 

\noindent{\textbf{Explaining deep learning models in healthcare}} 
\cite{lrp_hormone} used LRP
for explainability with an LSTM model on training dataset of 1048 patients 
and test dataset of 150 patients.
LRP backpropagates the prediction scores through the network
to provide explanations relative to input features. 
Initial confirmation was on digits images
from MNIST dataset, and global feature importance  
evaluated on the 15 test patients and on a per-patient 
basis on two patients. Relevant features were 
visually inspected and determined to be in alignment with clinical guidelines. 
To identify optimal refractive surgery technique,
\cite{shap_lasix} combined tree-based XGBoost classification model and
SHapley Additive ex-Planations (SHAP) to identify key factors used by
expert surgeons
on 18,480 Korean patients on non-sequential data, and achieved high degree of clinician-model agreement.  
The authors acknowledged the need for a validation scheme, and that objective
validation has yet to be found. We believe that our strategy of 
creating synthetic datasets which heavily mimics medical claims dataset,
with well-understood factors driving the labels will help address this gap.
Attention-based LSTM models were also briefly explored by \cite{att_icu} in predicting
clinical events 
in the ICU setting over a 14-day time window, and attention maps were used 
to provide global explainability to clinicians. The 229 input features utilized
were primarily of non-sequential nature from MIMIC-III dataset
and it is unclear which type of attention-mechanism was utilized. 

\noindent{\textbf{Attention for explainability in NLP}} 
\cite{att_is_not_explaination}
demonstrated that contrary to previous understanding, the 
attention weights were only weakly correlated to a number of
feature importance measures despite excellent prediction performance using
a test metrics such as F1-score and accuracy. 

\noindent{\textbf{Applying time decay toward patient temporal event sequence prediction}} 
\cite{context_embedding} introduced time decay factors applied on a diagnosis predictor that was based on similarities
between projected patient medical event sequences vectors and diagnosis vectors. 
The time decay factors were used to reflect the fact the more recent events have a bigger impact on the prediction. 

\vspace{-8pt}
\section{Data description}
\vspace{-5pt}
Significant time and resources are needed to generate ground truth labels, and 
domain experts often do not fully agree on causes of medical events 
for the patients in the real world. This leads to ambiguity that cannot be resolved when evaluating model explainability. To this end, we created simplified synthetic 
datasets
 that mimic medical claims datasets
while maintaining sufficient ground truth understanding to evaluate model explainability. We also utilize a real-world medical claims dataset for verification.
In all datasets, we seek to predict an adverse binary outcome. 
\vspace{-8pt}
\subsection{Synthetic dataset generation} 
\label{syn_d}
\vspace{-5pt}
Each patient is assigned a series of events, with events as proxies
of real diagnosis, procedure and medication billing codes. 
Events fall into four categories: adverse (A), helper (H), unhelper (U) and noise (N), 
and we impose fixed relationships for the impact of each category with 
respect to outcomes. Adverse (A) and helper (H) events 
represent severe disease diagnoses and
lead to higher probabilities of a bad outcome or positive label. 
Unhelper (U) events symbolize low risk, rewarding procedures that 
decrease the likelihood of adverse outcome. Noise(N) events 
represent insignificant, temporary side effects and 
does not impact the final outcome. There are 10 events (or tokens) each in
the A, H, U categories and 15 events for N category. 
Training dataset is balanced label-wise and equally from all possible
token combinations, with 21K observations generated. Validation and
test set contains 7K unique observations each. The highest frequency tokens
are from the N category, with each token occurring at about 6\% of all tokens;
(A, U, H) tokens occur at less than 1\% frequency.

\noindent{\textbf{Event-driven dataset:}} The presence or absence of
a set of events or tokens, without regard to the sequence of events, is the primary driver 
of binary outcome labels. This would be akin to
a set of events presented upon emergency room or intensive care unit admission,
where tests and alleviating actions are undertaken almost simultaneously.
Commonly, there are multiple causes that drive toward the same 
outcome and this understanding is incorporated into the dataset design. 
The encoded information is summarized in
Table \ref{tab:events}: seven combinations from 4 token categories with
45 total unique tokens, and results in 
different probabilities of a positive label. Random choice assigns labels 
for a given sequence using that probability threshold. Total sequence length is 
also random and capped at maximum 30 events.
We derive inspiration to create input tokens from cardiovascular-related events, 
and Table \ref{tab:events} provides the specific category combination,
probability of label and combination of events that should be accurately
captured by any ML model for explainability. For instance, 
sequence containing 2 adverse tokens and 1 helper token would 
have a 90\% chance of being assigned a positive label. Example of positive labels could be the 
occurrence of 30-day hospital re-admissions or all-cause mortality within a year.

\begin{table}[h]
\small
\centering
\begin{tabular}{|l|c|l|}
\hline
\begin{tabular}[c]{@{}l@{}}\textbf{Pathway or }\\ \textbf{token combination }\end{tabular} & \begin{tabular}[c]{@{}c@{}} \textbf{Probability} \end{tabular} & \textbf{Example} \\ \hline
2 adverse + 1 helper                                                    & 90\%                                                                     & \begin{tabular}[c]{@{}l@{}}myocardial infarction (A) + ventricular \\ hypertropy (A) + sleep apnea (H)\end{tabular} \\ \hline
1 adverse + 2 helper                                                    & 80\%                                                                     & \begin{tabular}[c]{@{}l@{}}pulmonary hypertension (A) + alchoholism (H) \\ + sleep apnea (H)\end{tabular}            \\ \hline
1 adverse + 1 helper                                                    & 70\%                                                                     & myocardial infarction (A) + pneumonia (H)                                                                            \\ \hline
1 helper + 1 unhelper                                                   & 40\%                                                                     & pneumonia (H) + diuretics (U)                                                                                        \\ \hline
1 adverse + 2 unhelper                                                  & 30\%                                                                     & \begin{tabular}[c]{@{}l@{}}pulmonary hypertension (A) + diuretics (U) \\ + ARBs (U)\end{tabular}                     \\ \hline
1 helper + 2 unhelper                                                   & 20\%                                                                     & sleep apnea (H) + MRAs (U) + beta blockers (U)                                                                       \\ \hline
2 unhelper                                                              & 10\%                                                                     & beta blockers (U) + ACE inhibitors (U)                                                                               \\ \hline
\end{tabular}\vspace{-8pt}
\caption{Combination of categories resulting in different likelihoods of 
positive label 
 to generate
the synthetic event-driven dataset. For sequence-driven dataset, these probabilities are 
additionally adjusted to account for time dependencies.}
\label{tab:events}
\end{table}\vspace{-8pt}

\vspace{8pt}
\noindent{\textbf{Sequence-driven dataset -- temporal effect of events:}}
Other applications, including by insurers, revolve around 
identifying patients likely to experience medical complications upon discharge, or
high-risk and high-cost adverse events for intervention. 
These datasets are typically have a temporal effect, e.g. more recent events have 
a bigger impact on the outcome in the near future, and that an ideal model 
should be able to capture the temporal context information 
for each medical event. We derive inspiration from \cite{context_embedding},
with the 
ground truth probability P of an adverse outcome for each dummy patient 
in the sequence dataset as:
\begin{multline}
P(O_t|A_{t_a,1},...,A_{t_a,m}, U_{t_u,1}, ..., U_{t_u,n},H_{t_h,1}, ... H_{t_h,p},N_{t_1},...,N_{t_q}) = \\
max(0.1,min(1, \sum e^{-a(t-t_a)} + \sum e^{-h(t-t_h)} - \sum e^{-u(t-t_u)}))   
\label{eq:1}
\end{multline}

where $$ 0<a<h<u<1 $$  $$t_a,t_h,t_u,t_n < t \le 0 $$

\noindent{$P(O_t| ... )$ represents the probability 
of an adverse outcome $P(O)$ at inference time (\textit{t}) given the 
number of adverse events (A) experienced by the patient at times $t_a$, 
unhelper events (U) at times $t_u$ and helper events (H) at times $t_h$, and 
one or more noise events (N) at multiple times ($t_1$ to $t_n$).
The term $e^{-\{a,h,u\}(t-t\{_a,_h,_u\})}$ is introduced to account for 
the decay of the impact of medical histories as discussed above. $(t-t\{_a,_h,_u\})$ is the time gap between the inference time and event time.
Variable \textit{a} is set to be a smaller positive constant because adverse events are causing 
higher likelihood of an adverse outcome. The unhelper (U) events 
have the minimal but beneficial impact on the patient’s outcome, 
and therefore the negative sign and the largest constant u. 
Notation-wise, current time step is set to be zero and previous times 
are set to be negative, i.e., in the past. }

\begin{wrapfigure}[15]{R}{0.6\textwidth} 
    \vspace{-5pt}
    \centering
    \includegraphics[width=0.9\linewidth]{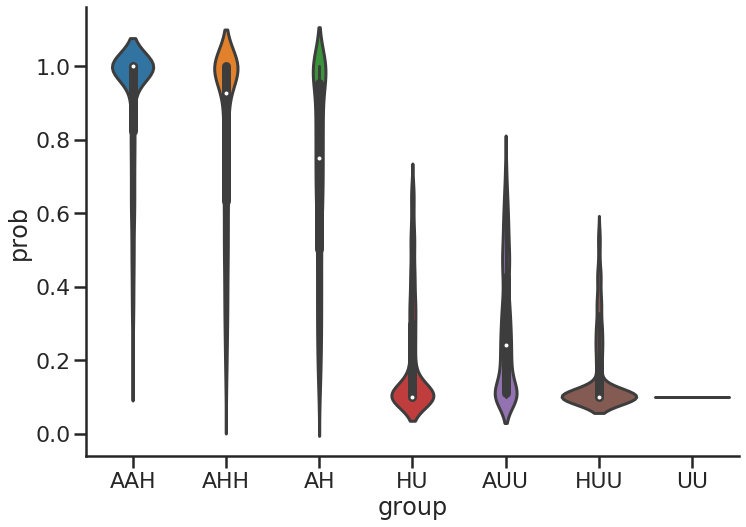}\vspace{-10pt}   
    \caption{Probability distribution for different pathways in sequence dataset}%
    \label{fig:freq_hist}%
\end{wrapfigure}

The average probabilities remain almost identical to Table \ref{tab:events}, but are
strongly adjusted (crossing the 50\% threshold) depending on the
relative ordering of the events. For instance, two adverse events followed by
a unhelper event (AAU sequence) will have lower probabilities relative to UAA sequence.
The further away these events have occurred from the time of prediction (\textit{t}=0),
that is the further in the past, the less likely a positive label will occur.
The temporal-adjusted probability distribution for each of the pathways
is depicted in Figure \ref{fig:freq_hist}.

\subsection{Real-world medical claims dataset} 
\label{cms}
The data is a 5\% sample limited dataset of the entire \cite{cms_dataset}
(CMS) medicare fee for service (Part A and B) claims data from 2009 to 2012,
with about 3 million unique beneficiaries 
containing information about medical events and claims such as diagnosis, procedures, admissions, 
discharges, and death. Standardized coding systems 
such as ICD-9, HCPCS, and CPT are used, 
and extracted for each patient from all healthcare settings, i.e. inpatient, outpatient, if available and consolidated. 
Other non-codified events including
admissions, discharges and gaps in medical care were provided unique tokens.

The events were codified with different prefixes depending on type. ICD-9 diagnosis events were coded in the format of "d\_" followed by the code; ICD-9 procedures as "p\_" and HCPCS procedures as "h\_". Gaps in medical care were codified in the format of "N\_days", where N refers to the numbers of days 
without any medical event activity between the last and next event.
The events were ordered chronologically with same day events ordered randomly.
Target event of interest is the Clostidium Difficile (\textit{C.Diff}) infection and 
coded via ICD-9 diagnosis code of 00.845. 
Our choice of \textit{C.Diff} infection event is due to the clinical significance and urgency of reducing in-hospital 
\textit{C.Diff} infections as jointly stated by the \cite{cdc_reduction} and CMS.

We chose to forgo elaborate 
feature engineering to philosophically mimic the experimental design of 
many prior deep learning approaches, including those previously mentioned approaches from \cite{rajkomar}, \cite{zhang}, and \cite{medbert}. All beneficiaries who had a primary hospitalization admission have at least one medical event are included. 
We additionally identified a list of potential risk factors (events) that lead to or cause \textit{C.Diff} infection, and 
this was achieved by curating existing literature 
and clinicians inputs. Similar to the synthetic dataset driver events, we used the risk factors as our ground truth. 
Finally, we include all medical events identified as risk factors and events having at least 
1000 occurrences in the dataset leading to a vocabulary of about 1000 medical events. 
Almost half of the selected events are the risk factors. The train set contains 1.4M observations while the validation and test sets contain around 248K observations each.



\vspace{-5pt}
\section{Methods}
\vspace{-5pt}
There are three degrees of freedom in this experiment: model, dataset and
explainability method. We explore a traditional ML model (XGBoost) and a 
deep learning model (LSTM); an event-driven, sequence-driven, and real-world dataset;
and for explainability: SHAP (both models), LRP(LSTM) and Attention (LSTM).
For synthetic dataset, we explore both simple dot-product attention and self-attention and only dot-product
attention for  real-world  dataset. For explainability, we quantify both global (overall feature importance) and
local (per patient feature importance).


\vspace{-5pt}
\subsection{Data preprocessing} 
\vspace{-5pt}
Different data preprocessing steps were applied depending on the target model. 
\noindent{\textbf{LSTM}} models can ingest variable sequences of events. The unique tokens form a vocabulary, 
with each token is assigned a unique index
number. This sequence of indices are directly passed in batches to LSTM models for training and inference. 
\noindent{\textbf{XGBoost}} models cannot ingest sequences of variable lengths.
Feature space is determined by the unique number of tokens in the entire dataset. And then each sequence is mapped to the feature space with the value in each dimension being the sum of occurrences of that token. Thus, all tokens or inputs are always
made available to the model, with 0 encoding used to denote the absence of a token in the sequence. 
The \noindent{\textbf{real-world dataset}}  requires additional preprocessing. 
For patients who received at least 1 medical event and had a primary hospitalization admission, 
up to 30 day's worth of medical events were captured. The choice of 30 day's was selected because longer history of data did not improve model performance. The events were captured relative to a 30-day 
forecasting window of interest. For example, if the forecasting window was January 1 2011, then the 
events prior to that date are captured. Additionally, the data is highly imbalanced with only less than 1\% positive examples. 
We used stratified sampling to ensure representative distribution of positive examples in validation and test sets, 
and down-sampled observations with negative labels in training dataset.



\subsection{Models, Methods and Metrics}
\vspace{-5pt}
\noindent{\textbf{Models}} Gradient boosting utilizes an ensemble of weak models to solve supervised ML problems, with
gradient boosted trees using decision trees as weak learners. XGBoost (eXtreme Gradient Boosting) is a highly 
efficient and scalable implementation of the gradient boosted trees algorithm. 
%
Long Short-Term Memory (LSTM) networks are a type of recurrent neural networks (RNNs) 
capable of learning from long sequences of data. Unlike feed-forward neural networks, 
RNNs utilize internal hidden states to represent information from previous events in the sequence.
Attention mechanisms are often added to the LSTM model architecture, which assists the model in selectively attending to
specific events in the input sequence. We utilize the simplest dot-product attention 
 and self-attention, and normalized attention scores are computed for each tokens in the sequence.

\noindent{\textbf{Training and Validation of Models}} 
Final models are decided using best AUC on validation set, which is the 
current widely used approach. We show later that this may not be the optimal approach
for explainability, and that explainability performance during training
as part of stopping criterion is also important. For speed purposes, a subset of 128-350 observations 
from the validation dataset is used to compute ground set truth similarities using the
iterative SHAP and LRP explainability approaches but full validation and test sets are evaluated with
the final models for AUC and explainability performance in synthetic datasets. For CMS dataset, we
evaluated explainability on all positive observations (160 in test set) and a down-sampling of negative observations for a total of 800 observations due to compute time. 

\noindent{\textbf{SHapley Additive exPlanations (SHAP)}} was developed by \cite{lundberg2017unified} based on a general idea of 
approximating the original complex model with a simpler explanation model. 
SHAP is categorized as an additive feature attribution methods, as the explanation model is
linear and can easily tell the effect of each feature for each prediction. The classic Shapley value is 
derived using classic equations from cooperative game theory, which is a weighted average of the all 
possible marginal differences of model output by adding the presence of the examined feature to 
subsets of features. We use a background of 300 
observations from training dataset to compute SHAP values. 

\noindent{\textbf{Layer-Wise Relevance Propagation (LRP)}} is originally developed to relate the input images pixels to the final prediction of the neural network. The concept is expanded into multiple domains which follow the general form that the classifier can be decomposed into several layers of computation. Such decomposition can be either achieved through relevance score back propagation or Taylor expansion. Since the model output can be approximated with a linear function of non-linear mappings of the original input through neuron activation, it can be interpreted as an additive feature attribution method as well.

\noindent{\textbf{Attention scores}} Besides offering super-human performance on many language understanding tasks, attention-based methods are being used to estimate feature importance. \cite{san_feature_importance} showed that Self-Attention Network (SAN) was able to identify feature interactions on a much larger feature subsets compared to the commonly used methods. In our case, the normalized dot-attention scores inherent in the LSTM model are extracted on a per patient basis for local feature importance comparisons, and we additionally explore self-attention
models in the synthetic dataset.

\noindent{\textbf{Metrics: Set similarity to ground truth for local explainability}}
We evaluate the models' ability to provide explainability at the per patient level.
This comparison is performed on a set similarity basis, where we measure the overlap of 
top \textit{N} important features between the SHAP output and the \textit{N} ground truth 
drivers of these events. This set similarity metric is calculated on a per patient 
basis, and then averaged over the entire validation dataset. 
Set similarity = 1.0 means that all the important drivers
were correctly identified in every patient, while zero similarity 
indicates that none of the key event drivers were correctly identified. Note that the
ground truth set for XGBoost and LSTM models are different, as XGBoost includes
the entire 45 tokens while LSTM models only see the active sequence, i.e. tokens
that correspond to one in the XGBoost input space.
\noindent{\textbf{Predictive performance}} is evaluated using AUC metric. 

\vspace{-5pt}
\section{Results on Synthetic Experiments: Event-driven dataset}
\vspace{-5pt}
We summarize findings on predictive performance, global feature importance for
overall model explainability and local feature importance for per-patient
explainability. 

\smallskip
\noindent{\textbf{Prediction Performance}} As the event-driven dataset 
does not depend on the sequential ordering of tokens and
is only influenced by the presence or absence of a set of tokens, 
the modeling is relatively simple and the prediction performance (AUC) 
of both models are comparable. Best AUC results from
training, validation and test sets are shown for 
both models in Table \ref{tab:auc_perf} where AUC results reach beyond 0.8 using
this event-driven dataset. For
both models, the highest AUC epoch is evaluated for explainability.

\begin{table}[htbp]
\small
\centering
\begin{tabular}{|c|c|c|c|c|}
\hline
\textbf{Model} & \textbf{Dataset}                 & \textbf{Training AUC} & \textbf{Validation AUC} & \textbf{Test AUC} \\ \hline
XGBoost        & Event-driven                   & 0.8821                & 0.8174                  & 0.8177            \\ \cline{1-1}\cline{3-5}
LSTM           &                                  & 0.8112                & 0.8249                  & 0.8278            \\ \hline
XGBoost        & Sequence-driven       & 0.8970                & 0.8282                  & 0.8321            \\ \cline{1-1}\cline{3-5}
LSTM           &                                  & 0.8623                & 0.8706                  & 0.8763            \\ \hline
\end{tabular}\vspace{-5pt}
\caption{Best AUC performance by both models on training, validation and test sets for datasets with and without temporality.}
\label{tab:auc_perf}
\end{table}\vspace{-15pt}





\vspace{12pt}
\smallskip
\noindent{\textbf{Global explainability}} 
Majority of negative labels contains unhelper (U) tokens (see Table \ref{tab:events}) compared to 
those with positive labels due to the designed pathways, and these tokens are expected to have 
the highest or close to highest importance in the ranking of global features. 
Adverse (A) tokens should have similar importance or follow very closely next in ranking, as they are predictors of the
positive labels but can also be reduced in impact by the presence of unhelper (U) tokens (30\% probability of positive label when
paired with 2 Us). Helper (H) tokens should be a distant third ranked group, as they are present in 
majority of the pathways while noise (N) tokens will be last due their lack of predictive power. 
In summary, the order of set of tokens should be a mix of (U's, A's), H's, and finally
N's (non-information containing).

\begin{figure}[!ht]
    \vspace{-10pt}
    \centering
    \includegraphics[angle=270, width=0.8\textwidth, keepaspectratio]{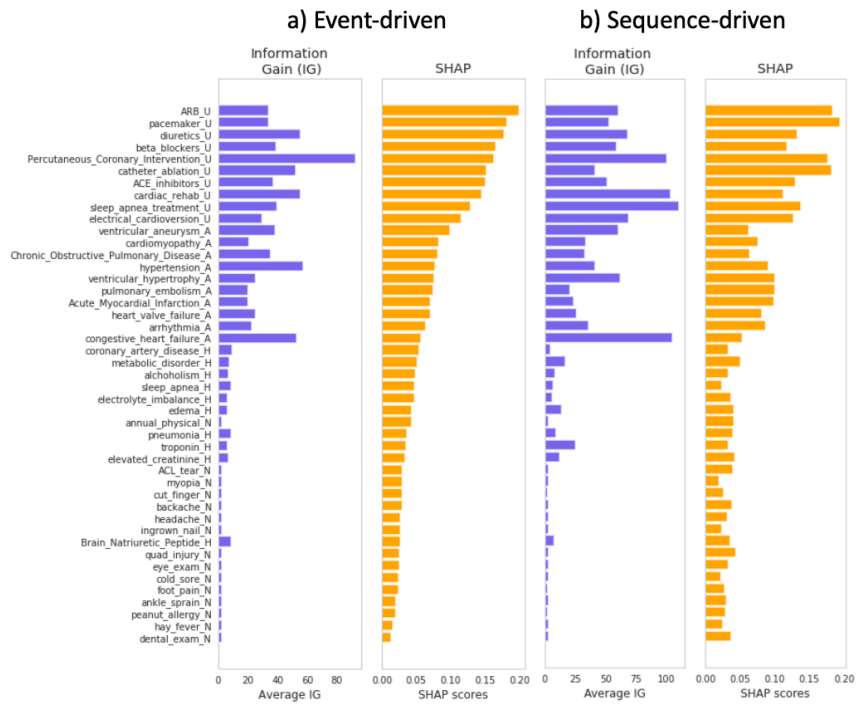}\vspace{-5pt}
    \caption{Global feature importance using IG and SHAP with XGBoost models on a) event-driven dataset and
    b) sequence-driven dataset, both using validation sets.}
    \label{fig:global_ig_xgb}
\end{figure}

For the XGBoost model, the relative influence of the 45 tokens 
is well aligned with the known event drivers. Figure \ref{fig:global_ig_xgb}a) summarizes
the global importance calculated using absolute SHAP scores, and plotted against 
global feature importance based on average information gain (IG).
As with SHAP, the IG token classes were generally ordered correctly but the exact
ordering between the tokens differ. 
The demarcation between information-containing and noise tokens is not clear
with SHAP, but a cutoff can be fairly easily applied via IG.

\begin{figure}[!ht]
    \centering\vspace{-20pt}
    \includegraphics[angle=270, width=0.7\textwidth, keepaspectratio]{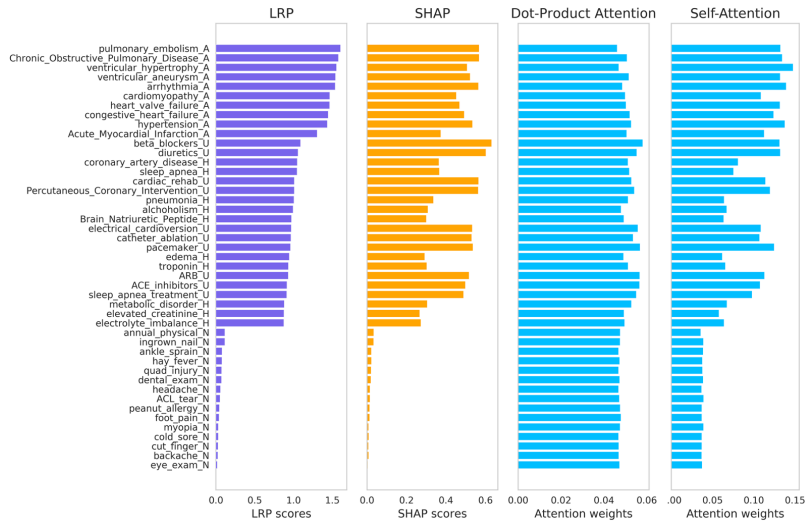}\vspace{-20pt}
    \caption{Global feature importance using LRP, SHAP, Dot-Attention, and Self-Attention with LSTM models on event-driven validation dataset.}
    \label{fig:global_lstm}
\end{figure}
\begin{figure}
    \centering
    \vspace{-60pt}
    \hspace{-10pt}
    \includegraphics[angle=270, width=\textwidth]{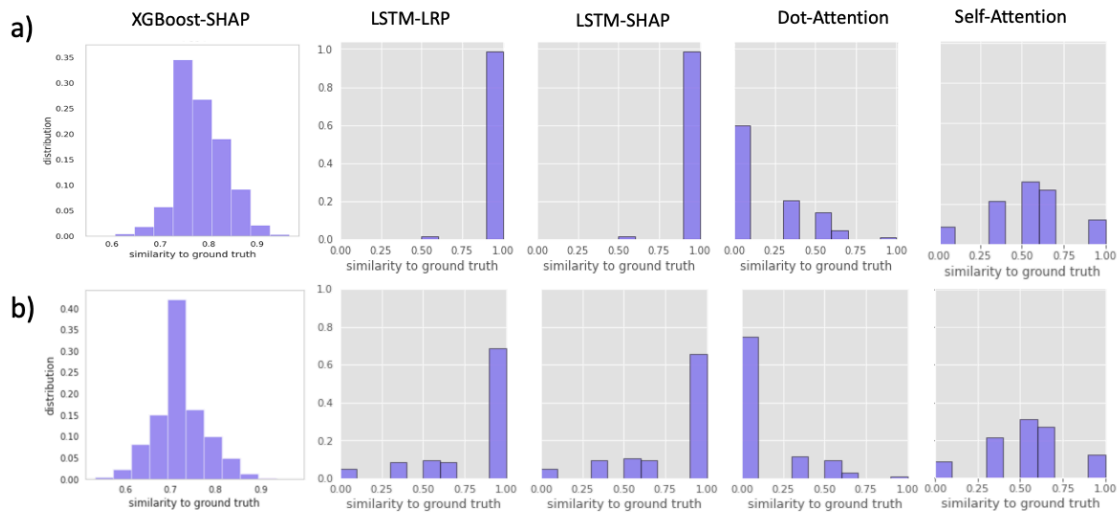}    
    \vspace{-70pt}
    \hspace{-10pt}
    \caption{Set similarity distribution of the top \textit{N} tokens from XGBoost, LSTM versus designed ground truth tokens, on a) event-driven and b) sequence-driven test set}
    \label{fig:sim_dist}%
\end{figure}%



For the LSTM model, the four different approaches (SHAP, LRP, Dot-Product Attention and Self-Attention) 
to global model explainability is shown in Figure \ref{fig:global_lstm}. 
For LRP, the set of adverse (A) tokens dominate explainability, 
followed by intermixing of unhelper (U) and helper (H) tokens.
Noise tokens are correctly attributed with negligible importance. For explainability by SHAP values, the most important
tokens are a mix of (A, U) tokens followed by helper tokens, and also assigns correctly little value to noise tokens. 
Dot-product attention's feature importance is almost uniform across all tokens, and does not provide
meaningful discrimination. Self attention provides better discrimination between useful and noise tokens than 
dot-product attention.

\smallskip
\noindent{\textbf{Local explainability}} The statistics (mean, range) applied to the test dataset 
for set similarity scores is reported in Table \ref{tab:set_scores}. Unlike LSTM, every feature or token 
is considered per observation with XGBoost and is compared
against the existing 30 information-containing tokens. The distribution of set similarity 
scores per patient for LSTM is much more discrete than XGBoost,
since there is at most 3 driver tokens at any sequence. Both LRP and SHAP are able to identify
the information-containing tokens from the noise tokens very well, while attention scores
often failed to do so. The score distribution is shown in Figure \ref{fig:sim_dist}a).

\begin{table}[!ht]
\small
\centering
\begin{tabular}{|c|c|c|c|c|c|c|}
\hline
\textbf{Metric} & \textbf{Dataset} & \textbf{\begin{tabular}[c]{@{}c@{}}XGBoost\\ SHAP\end{tabular}} & \textbf{\begin{tabular}[c]{@{}c@{}}LSTM\\ SHAP\end{tabular}} & \textbf{\begin{tabular}[c]{@{}c@{}}LSTM\\ LRP\end{tabular}} & \textbf{\begin{tabular}[c]{@{}c@{}}LSTM\\ DotAttention\end{tabular}} & \textbf{\begin{tabular}[c]{@{}c@{}}LSTM\\ SelfAttention\end{tabular}} \\ \hline
Mean & \multirow{2}{*}{\begin{tabular}[c]{@{}c@{}}Event-\\ driven\end{tabular}} & 0.79 & 0.993 & 0.993 & 0.178 & 0.85 \\ \cline{1-1} \cline{3-7} 
Range &  & {[}0.567, 0.933{]} & {[}0, 1{]} & {[}0, 1{]} & {[}0, 1{]} & {[}0, 1{]} \\ \hline
Mean & \multirow{2}{*}{\begin{tabular}[c]{@{}c@{}}Sequence-\\ driven\end{tabular}} & 0.723 & 0.805 & 0.818 & 0.117 & 0.526 \\ \cline{1-1} \cline{3-7} 
Range &  & {[}0.533, 0.933{]} & {[}0, 1{]} & {[}0, 1{]} & {[}0, 1{]} & {[}0, 1{]} \\ \hline
\end{tabular}\vspace{-10pt}
\caption{Local explainability: set similarity scores for each method on both test datasets (event-driven, sequence-driven).}
\label{tab:set_scores}
\end{table}\vspace{-10pt}

%



\section{Synthetic dataset: Sequence-driven dataset}
\vspace{-5pt}
\noindent{\textbf{Prediction Performance}} Due to the
sequential nature, whereby the relative ordering of the tokens and
recentness of tokens affect the label probability, we 
observe performance differences between XGBoost and LSTM in the sequence-dataset section of Table \ref{tab:auc_perf}.


\noindent{\textbf{Global explainability}} 
The added time dependencies results in a significant spread in label probabilities 
used to create the data (Figure \ref{fig:freq_hist}).
It is expected that the adverse (A) and unhelper (U) events to dominate, but with more intermixing due to the time-dependencies.
The models should be able to separate noise from other informative tokens.

\begin{figure}[!ht]
    \vspace{-30pt}
    \centering
    \includegraphics[angle=270, width=\textwidth, keepaspectratio]{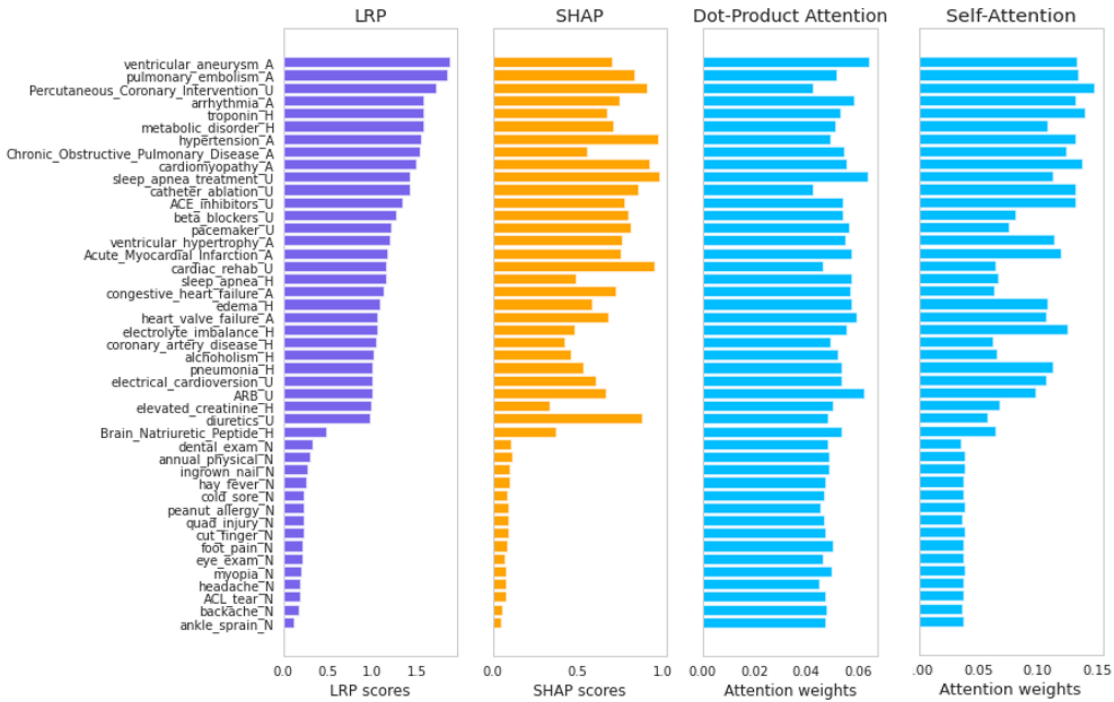}\vspace{-25pt}
    \caption{Comparison between global feature importance using LRP, SHAP, Dot-Attention and Self-Attention with LSTM models for sequence-driven dataset.}\vspace{-10pt}
    \label{fig:global_lstm_seq}
\end{figure}

For the XGBoost model, the relative influence of the adverse and unhelper tokens are correctly identified but
struggles to separate the helper tokens from the noise. Figure \ref{fig:global_ig_xgb}b) summarizes
the global importance via SHAP and IG for sequence-driven dataset. For the LSTM model, 
global model explainability via LRP and SHAP continue to 
be accurate, and is shown in Figure \ref{fig:global_lstm_seq}. There is significant
intermixing of the tokens as expected but the demarcation between noise and informative tokens is clear.
Dot-product attention's feature importance does not provide
meaningful discrimination, with self-attention displaying discriminative ability
between information and noise tokens, but with less clear demarcations compared to LRP or SHAP.

\smallskip
\noindent{\textbf{Local explainability}} 
Local explainability performance on the test set is summarized in the bottom half of Table \ref{tab:set_scores}. 
LSTM's explainability performance via SHAP and LRP are higher than XGBoost. Distributions of this
per patient explainability performance is shown in Figure \ref{fig:sim_dist}b),
where a majority of the explanations per patient was correct, and skewed by a minority of
observations for LSTMs with SHAP or LRP. Self-attention continues to outperform dot-product attention.

\begin{figure}[!ht]
    \vspace{-50pt}
    \centering
    \includegraphics[height=0.6\textheight, width=0.95\textwidth]{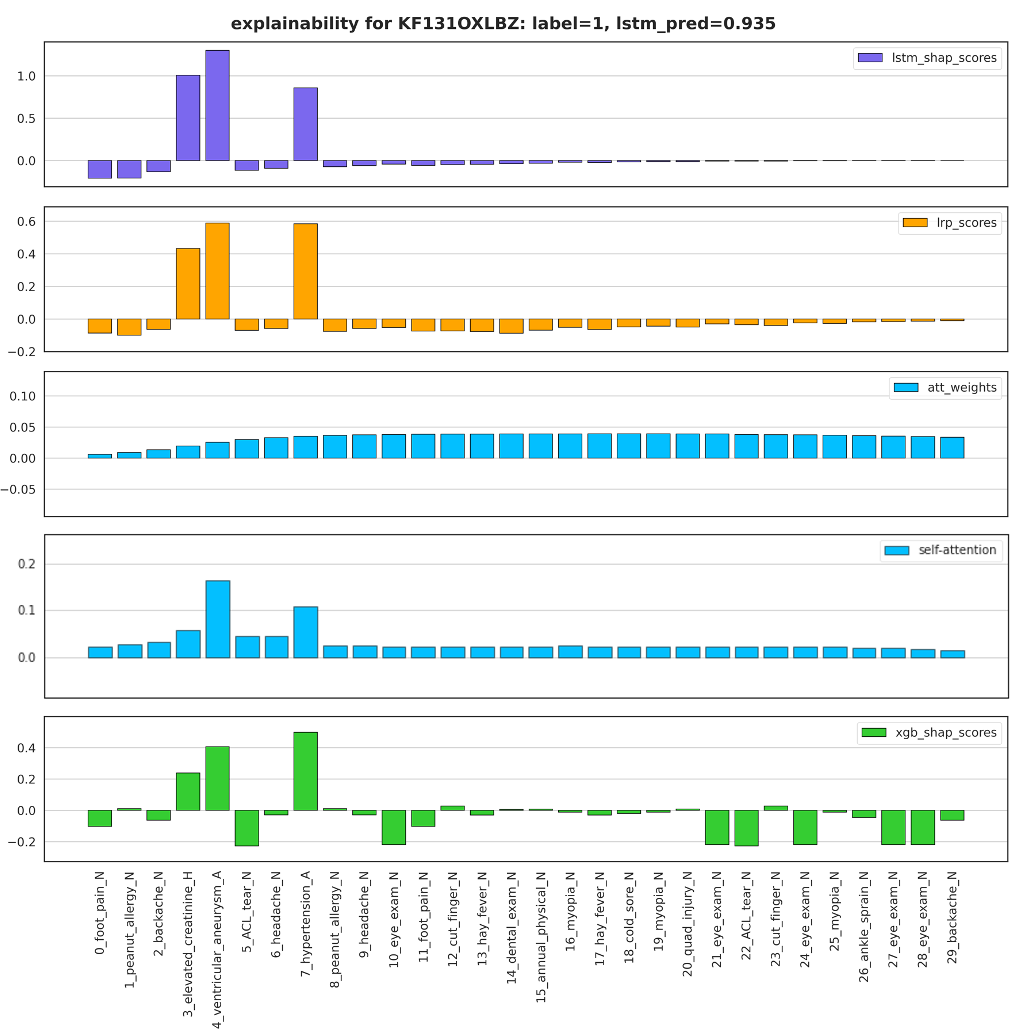}
    \includegraphics[height=0.35\textheight, width=0.95\textwidth]{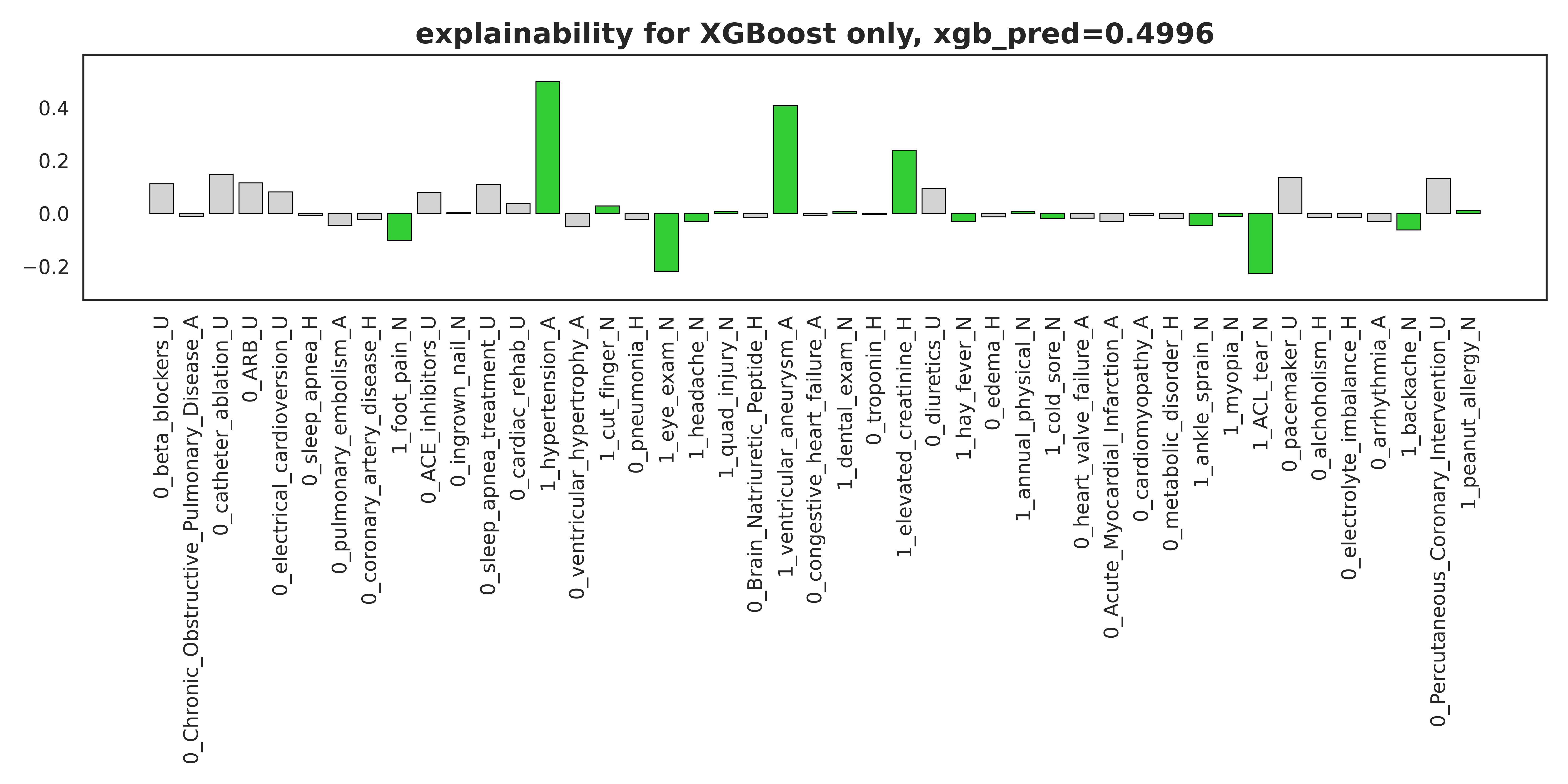}
    \caption{Example where XGBoost and LSTM models disagree on local explainability due to the temporal nature of the events.}
    \label{fig:diff_example}
\end{figure}

A full example of sequences with local explainability on sequence-driven data is shown in Figure \ref{fig:diff_example}.
The top five figures correspond to the actual sequence of tokens or events and the relative importance scores 
assigned by LSTM-SHAP, LSTM-LRP, LSTM-Dot-Attention, LSTM-Self-Attention and XGBoost-SHAP. All the importance tokens were correctly
identified. The x-axis denotes the combination of (sequence index, token, category), with \textit{0\_foot\_pain\_N}
being the first token of event \textit{foot\_pain} from the noise (N) category. The lowest figure denotes the full input
feature space into the XGBoost model, which includes absence of tokens and highlights the inherently different
input feature space encountered by both models. Positive events in green
correspond to the earlier plot, which only displayed events that occurred. Some SHAP values were 
attributed to the non-occurring tokens, indicating that the sequential-ness of the tokens affect XGBoost's ability to predict correctly, 
and noise tokens such as \textit{eye\_exam} were inaccurately attributed.

\vspace{-5pt}
\subsection{Evolution of explainability during model training}
\begin{figure}[!ht]
    \centering
    \vspace{-40pt}
    \includegraphics[angle=180,width=0.8\textwidth]{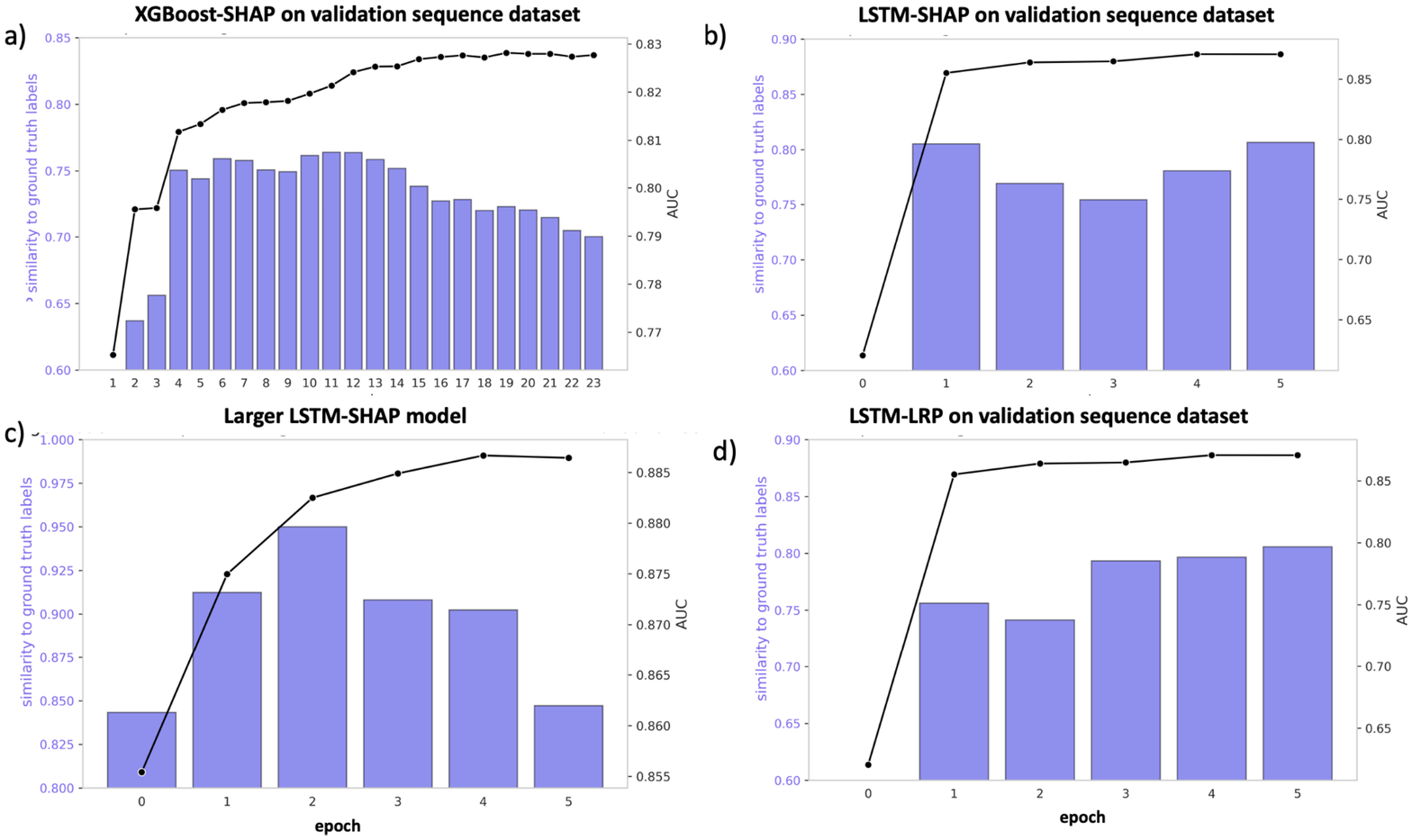}
    \vspace{-40pt}
    \caption{Evolution of predictive and explainable performance during model training on the
    sequence synthetic validation dataset. a) XGBoost-SHAP validation performance and AUC prediction performance. 
    b) LSTM-LRP performance on a right-sized model and learning rate, c) LSTM-LRP performance on a slightly
    too large model and learning rate, and d) LSTM-SHAP performance on right-sized model and learning rate}
    \label{fig:evolution}
\end{figure}

\vspace{-5pt}
For XGBoost model during training on both synthetic datasets, we noticed a trend of 
improved explainability performance followed by a decrease. For the sequence-driven dataset in Figure \ref{fig:evolution}a, 
Epoch 11 has the best explainability performance
among all epochs with a set similarity of 0.764 on validation set. 
The highest AUC would be in epoch 19, which provided
a reduced set similarity score of 0.723 or roughly 5\% reduction in explainability. 
Visually, the optimal trade-off would be around Epoch 12. 

We did not observe similar trends for the LSTM model for the set of model parameters and learning rates
used (hidden state dimension = 16, embedding dimension = 8, and slow learning rate = 0.0003) for model training,
which is shown in Figure \ref{fig:evolution}b for SHAP explainability and Figure \ref{fig:evolution}d for
LRP explainability. These values were arrived using recommendations for model training listed in 
discussion section for optimal predictive and explainability performance, and we note a high sensitivity of
explainability on right-sizing the model, regularization and learning rate. 
When the model parameter space and learning rates 
are slightly too large, we do observe similar trends to that displayed by XGBoost. This is shown in Figure 
\ref{fig:evolution}c, using hidden state dimension = 16, embedding dimension = \textbf{16}, and 
learning rate = 0.001 (about 3x compared to ideal learning rate). 

Additional reduction in learning rates and increases in regularization did not remove this trend for XGBoost. 
This serves to emphasize the importance of having some understanding of important tokens or features within
the input space, and to incorporate the set similarity overlap between these known features and important
model features during model training. 

\vspace{-5pt}
\section{Results on Real Data: \textit{C.Diff} prediction} 
\vspace{-5pt}
\noindent{\textbf{Prediction Performance}} Predictive performance
is summarized in Table \ref{tab:auc_perf_cms}.

\begin{table}[htbp]
\small
\centering
\begin{tabular}{|c|c|c|c|}
\hline
\textbf{Model} & \textbf{Training AUC} & \textbf{Validation AUC} & \textbf{Test AUC} \\ \hline
XGBoost        & 0.9170               & 0.8772                  & 0.9184         \\ \hline
LSTM           & 0.9292                & 0.8932                 & 0.9226            \\ \hline
\end{tabular}
\caption{Best AUC performance by both models on training, validation and test sets}
\label{tab:auc_perf_cms}
\end{table}

\begin{figure}[!ht]
    \vspace{-10pt}
    \centering
    \includegraphics[width=0.65\textwidth, height=0.3\textheight]{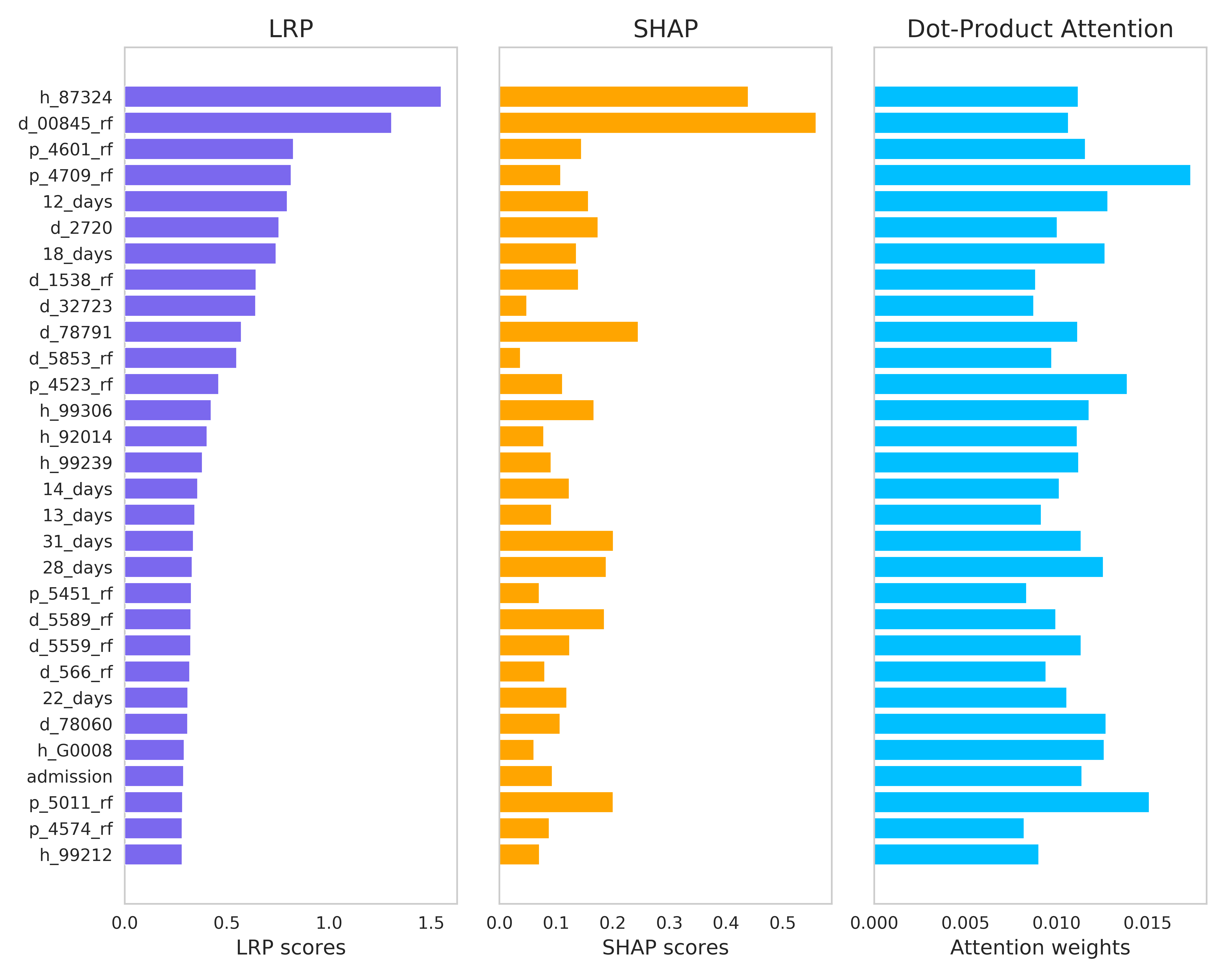}\vspace{-8pt}
    \caption{Comparison between global feature importance (top 30 scores) using LRP, SHAP and Dot-Attention with LSTM models for \textit{C.Diff} dataset.}\vspace{-5pt}
    \label{fig:global_lstm_cdiff}
\end{figure}

\begin{wrapfigure}[19]{r}{0.55\textwidth}
    \centering
    \includegraphics[width=0.55\textwidth]{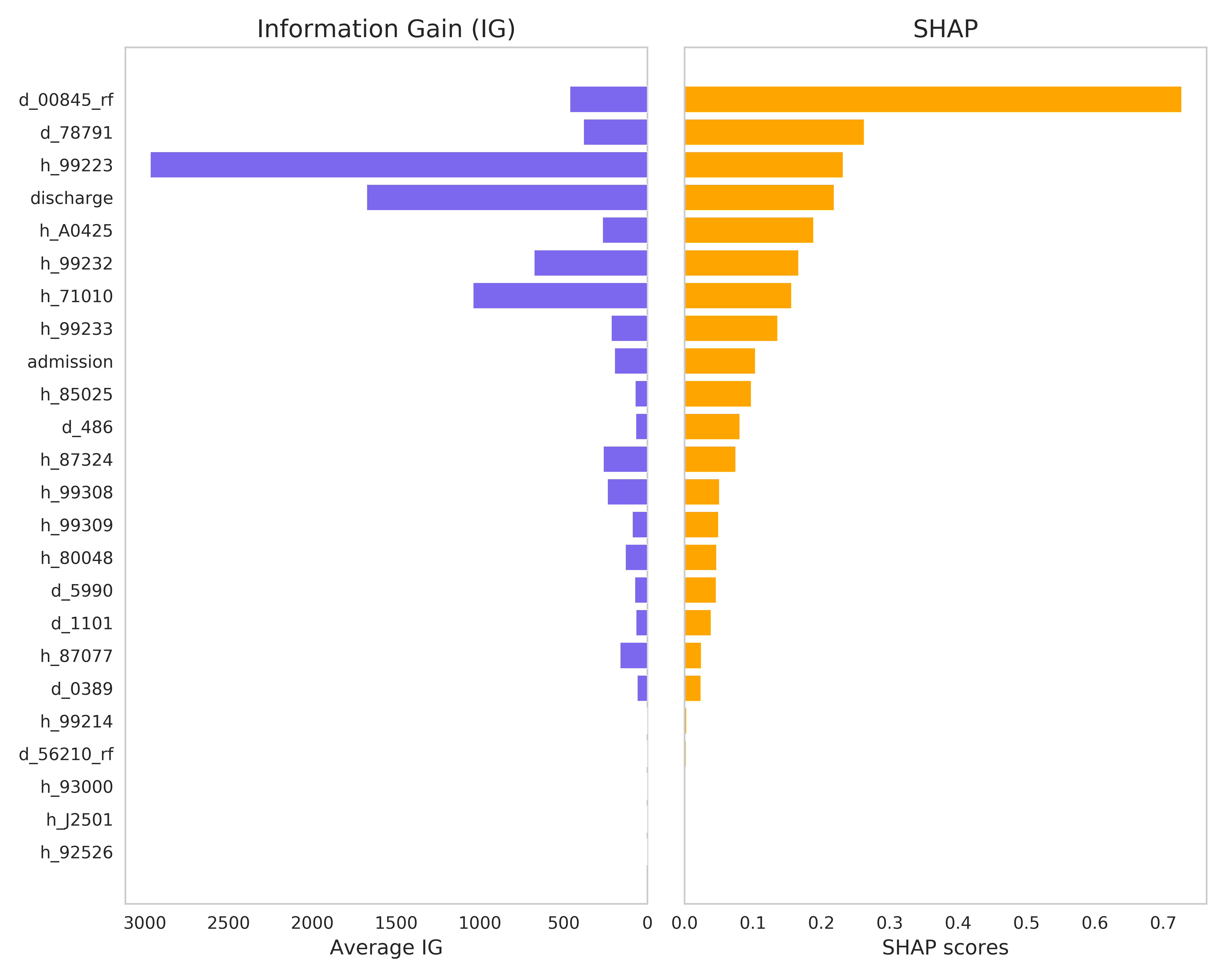}\vspace{-16pt}
    \caption{Global feature importance using IG and SHAP with XGBoost model on validation set for \textit{C.Diff} dataset.}
    \label{fig:global_ig_xgb_cdiff}
\end{wrapfigure}

\noindent{\textbf{Global Explainability}} The global feature importance from the LSTM model
is shown in Figure \ref{fig:global_lstm_cdiff},
using the best model identified over model training on validation data. Tokens that have suffixes of \textbf{\_rf} are
the identified risk factors from literature and clinicians. We find top 30 important tokens when predicting
\textit{C.Diff} infections by the model to be previously identified risk factors (12/30), gaps between times the patient receives medical care and
hospital admission. The global feature importance by XGBoost is shown in Figure \ref{fig:global_ig_xgb_cdiff}, and only 2 of the 24 top
features were associated with previously known risk factors.

Among those top identified risk factors from the LSTM model, we see the ICD-9 `d\_00845\_rf' diagnosis token, which refers to previous \textit{C.Diff} infection. The ICD-9 `p\_4601\_rf' procedure token refers to `Exteriorization of small intestine" and `p\_4709\_rf' refers to an appendectomy. The HCPCS code `h\_87324' was not identified as a risk factor, and is lab result of `Infectious agent antigen detection by immunoassay technique'. The combination of these highlighted token paints a global picture that previous infections of \textit{C.Diff}, along with bowel related procedures and confirmed lab tests were highly predictive of \textit{C.Diff} diagnosis. 

\begin{wrapfigure}[12]{R}{0.65\textwidth}
   \centering\vspace{-15pt}
   \includegraphics[width=0.65\textwidth, height=0.18\textheight]{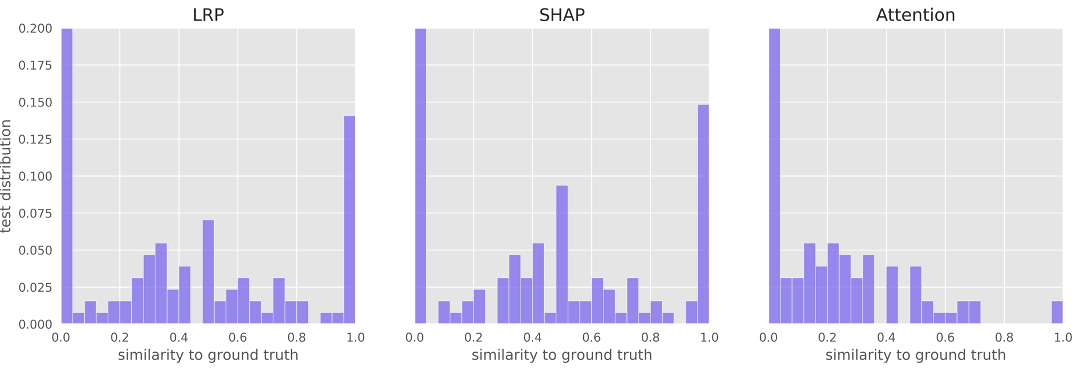}\vspace{-22pt}
   \caption{Distribution of the set similarity of the top \textit{N} tokens from 
        LSTM and the known ground truth tokens, on validation and test set on \textit{C.Diff} dataset.}
    \label{fig:lstm_sim_dist_cdiff}
\end{wrapfigure}

\smallskip
\noindent{\textbf{Local Explainability}} The set similarity metrics from SHAP, LRP and Dot-Attention with LSTM is shown in Figure \ref{fig:lstm_sim_dist_cdiff}.  The patients without any identified risk factors and are excluded from calculations
and distributions plots. The average set similarity for LRP, SHAP and Dot-Attention are 0.264, 0.305 and 0.017 and all with
ranges of [0.0, 1.0]. Similar trends to synthetic data is observed, where 
attention scores underperform relative to LSTM-LRP and LSTM-SHAP. The high frequency of 
zero similarities originate from the fact that a good portion of patients had only
one known risk factor and that the top feature identified by the model was not the single risk factor.
The set similarity mean and range for XGBoost-SHAP is 0.45 and 
[0.419, 0.502]. Due to the CMS Data Use Agreements, individual explainabiltiy outputs cannot be shown.

\section{Discussion} 
\vspace{-5pt}
\noindent{\textbf{Strongly prevent over-fitting for explainability}} It is standard practice to utilize a validation set to
determine optimal model parameters, and model with the highest predictive
performance is selected for further use. First, we have shown that
the model with highest \textit{explainability} performance may not be the model 
with the best \textit{predictive} performance. 
Thus, an optimal stopping point
should be evaluated using both predictive metrics such as AUC and 
against known important features per epoch.
Secondly, the model tuning for explainability should be performed
with care -- with continued reduction of parameter dimensions and increase of
regularization until noticeable degradation in predictive performance is observed. 
One can then conclude that the smallest parameter space has been found. This is particularly
true for deep learning models, where the number of parameters or weights to be trained
are often on the same order of magnitude as the available observations. We obtained similar
predictive performances using a just-right model and a too-large model
but with very different explainability performances, which may lead to wrong
clinical conclusions.

\smallskip
\noindent{\textbf{Dot-product attention is not explainability}} 
Attention scores do not provide good overall explainability on
global \& locally, 
although it does accurately identify information-containing tokens in a small
amount of examples and should not
be used to provide explainability in a clinical setting. 

\smallskip
\noindent{\textbf{Time-dependency of dataset}}
An important aspect to model selection is to determine whether temporal 
dependencies exist with respect to the
events driving the class label. When temporality exists or is uncertain, 
the LSTM model with SHAP or LRP explainability 
provides robustness in both 1) identification of global key features and 
2) local explainability that can be provided to convince clinicians of correct
model reasoning. 
\vspace{1pt}

\smallskip
\noindent{\textbf{Separation of key features and noise}}
For identification of global key features, the relatively sharp drop-off 
(Figure \ref{fig:global_lstm_seq})
in LSTM-SHAP and LSTM-LRP scores makes it relatively easy to distinguish
between important and non-important features using either method. This is
slightly less clear with XGBoost-SHAP alone, which captures majority of the
overall important features but has trouble distinguishing between noise tokens 
and helper tokens. This gradual taper of SHAP scores 
makes it difficult to determine information-containing features as
easily as the LSTM-SHAP/LRP combination. We note that if a traditional ML model must
be utilized, one can leverage the comparatively sharp cutoff provided IG to distinguish between features containing information
about the classification task and those that do not, but the 
better relative ordering between the tokens by SHAP indicate that the
relative importance between features are better represented by SHAP than IG.

\smallskip
\noindent{\textbf{Use LSTM-SHAP or LSTM-LRP for local explainability}}
The average set similarity between ground truth labels and model-identified features
are highest using LSTM-SHAP or LSTM-LRP. When the temporality of the features is 
non-existent or weak, XGBoost-SHAP has the added value of explicitly considering 
the \textit{absence} of a token as opposed to the implicit approach of LSTMs and may
be useful in certain contexts. The choice between SHAP or LRP should be based on required inference time. 
SHAP 
scales with the number of background examples applied for each prediction which increases inference time, but there exists 
libraries for easy use of SHAP. 
Conversely, LRP back-propagation is custom and specific to each model definition and architecture 
but only needs to calculate the score back-propagation once per observation. 
In instances such as urgent care setting or during operative procedures where real-time evaluations are needed, 
it may be advantageous to invest resources upfront via LRP customization in exchange for low inference time. For
other settings, such as classifications of patients for additional follow up care post-discharge, 
this difference in inference time does not make much difference.

\smallskip
\noindent{\textbf{Considerations for Real World Modeling}}
Compared to the synthetic dataset, the real world results showed similar explainability characteristics in that known risk factors were identified by LRP/SHAP and some demarcation between events can be seen. However without a complete ground truth, interpreting the "correctness" of the explainable outputs remains uncertain. Real world clinical ground truths can vary significantly depending on the clinician and presents challenges in optimizing model training for explainability. Known risk factors may not necessarily manifest depending on the individual and factors previously not considered relevant may later be discovered to be significant. This lack of a complete ground truth presents challenges on how to optimize a real world model for explainability.

Nonetheless, targeting good explainability remains important and we present several strategies to be explored to alleviate the challenges. If significant risk factors are unclear on the medical event level, then grouping risk factors and evaluating explainability on a class level could give directional feedback on its quality. It may also be easier to instead identify known noisy features and look for clear demarcation between the known noisy events and others. Smaller scale ground truth analysis from individual cases could also be validated through set similarity metrics looking at local explainability. Pertaining to the generalizability of LRP and SHAP, these methods can be ensembled, highlighting features universally agreed between different approaches. 

\paragraph{Limitations}
The input space we explored in the synthetic dataset was limited to 45 to
ensure accurate and easy ground-truth comparison. In contrast, 
thousands of medical codes are currently in active use. Further
work should be utilized to understand this in conjunction with the
degradation of explainability performance when there are 
too many model parameters, as large vocabularies lead to learning exceedingly large
number of embeddings. This is often addressed by grouping similar codes together 
using medical concepts.

The classification tasks in this study had relatively high AUC performances (above 0.8). It is unclear how our conclusions will change for lower performing models. We assume that explainability degrades with lower model performance, and further studies into the degree of degradation at different levels of performance are needed. We also focused on understanding the explainability performance of models on
sequence-based, medical claims-like datasets. It is unclear how this
would be extrapolated to other types of datasets like EHRs. Understanding derived from this work is extendable to simpler, 
less complex datasets.

This study does not include transformer models, which are best utilized 
with datasets larger than what was available in the real-world dataset. Existing transformer models are pre-trained on EHR datasets and cannot be leveraged for claims datasets. Finally, the multi-headed attention mechanism native to transformers does not fit the scope of our evaluation as they describe inter-feature relationships rather than feature-prediction relationships.


\bibliography{references}



\end{document}